\documentclass{vgtc}                          

\usepackage{color}
\usepackage{adjustbox}
\usepackage{rotating}
\usepackage{dblfloatfix}
\usepackage{booktabs}
\usepackage{color}
\usepackage{array}
\usepackage{mathtools}

\usepackage{amssymb}
\usepackage{amsmath}
\usepackage{float}

\usepackage{mathptmx}
\usepackage{graphicx}
\usepackage{times}
\usepackage{subcaption}
\onlineid{0}

\vgtccategory{Research}

\vgtcinsertpkg

\title{Multi-layer Visualization for Medical Mixed  Reality}

\author{S\'everine Habert \thanks{e-mail: severine.habert@tum.de}  \\
        \parbox{1.6in}{\scriptsize \centering Technische Universit\"at
        M\"unchen}
      \and Ma Meng\thanks{e-mail: meng@in.tum.de }\\ %
        \parbox{1.6in}{\scriptsize \centering Technische Universit\"at
        M\"unchen}
\and Pascal Fallavollita\thanks{e-mail: fallavol@in.tum.de}\\ %
  \parbox{1.6in}{\scriptsize \centering Technische Universit\"at M\"unchen} %
\and Nassir Navab\thanks{e-mail: navab@cs.tum.edu}\\ %
     \parbox{1.6in}{\scriptsize \centering Technische Universit\"at M\"unchen \\
     Johns Hopkins University}}

\abstract{ Medical Mixed Reality helps surgeons to contextualize
intraoperative data with video of the surgical scene. Nonetheless, the surgical
scene and anatomical target are often occluded by surgical instruments and
surgeon hands. In this paper and to our knowledge, we propose a multi-layer visualization in Medical Mixed Reality solution which subtly improves a surgeon’s visualization by making
transparent the occluding objects. As an example scenario, we use an augmented
reality C-arm fluoroscope device. A video image is created using a
volumetric-based image synthesization technique and stereo-RGBD cameras mounted
on the C-arm. From this synthesized view, the background which is occluded by
the surgical instruments and surgeon hands is recovered by modifying the
volumetric-based image synthesization technique. The occluding objects can
therefore become transparent over the surgical scene. Experimentation with
different augmented reality scenarios yield results demonstrating that the
background of the surgical scenes can be recovered with accuracy between
45\%-99\%. In conclusion, we presented a solution that a Mixed Reality solution for medicine, providing transparency to objects occluding the surgical scene. This
work is also the first application of volumetric field for Diminished Reality/
Mixed Reality.
} 


\begin{document}


\firstsection{Introduction}
\maketitle

The term ``Surgery'' comes from the Greek ``Kheirourgia" which means handiwork.
Despite numerous technological improvements in the last centuries, surgery
remains a manual work where surgeons perform complex tasks using their hands and
surgical instrumentation.
As it is yet not possible to retrieve the view as seen directly by the surgeon,
numerous works are using video cameras to record the entire surgical scene. Such
a solution is applicable for training medical students using ``first-person"
view cameras \cite{bizzotto2013video}, or more commonly for Medical Augmented
Reality where another modality (intraoperative or preoperative) is overlaid over
the video to give context to the medical data.
Having the hands and instrumentation positioned in the field of action
inherently signifies the occlusion of the surgical scene and the anatomy being
treated. This is true both from the surgeon viewpoint or any imaging modality
viewpoint. It would be advantageous if there was a solution to display to the
surgeon any occluded region of interest without losing the information about the
action that is given by the hands and surgical instrument positions. Introducing
transparency links the problem to the Diminished Reality field of study.
Such an application would then combine Diminished Reality with Augmented Reality, providing a Mixed Reality Visualization.

\subsection{Related Works}

Medical Augmented Reality can be classified into 2 main categories: preoperative
data (CT, MRI) overlaid over intraoperative data (video, X-ray images), or
intraoperative data overlaid over intraoperative data stemming from another
modality. The first category uses preoperative data to segment 3D models of
organs  \cite{onda2013short,liao20103} or plan paths/entry points
\cite{de2010augmented} that can then be rendered during surgeries using video
coming from an external camera or an endoscope. The second category uses
intraoperative data acquired during surgery to display over another type of
intraoperative data, most of the time video. The overlaid intraoperative data
can be 3D such as 3D Freehand SPECT images \cite{okur2011mr}, 2D such as X-ray
images \cite{navab2010camera}, OCT images \cite{roodaki2015introducing} or
ultrasound \cite{sauer2001augmented}.

The Camera Augmented Mobile C-arm by Navab et al. \cite{navab2010camera} has
been the first Augmented Reality device to enter an Operating Room and has been
used on over 40 patients  \cite{navab2012first}. A video camera is placed next
to the C-arm source and a mirror construction fixed under the X-ray source
allows the alignment of the optical axis and centers of both modalities such
that an exact overlay of X-ray and video is possible. The main drawback of this
work is its mirror construction, which restricts the surgical workspace
available for the surgeon and requires invasive engineering on the C-arm.
Habert et al. \cite{habert2015posteraugmenting} proposed to augment a C-arm with
2 RGBD cameras placed on the side of the X-ray source. Using the RGBD data, the
video image from the X-ray source viewpoint can be synthesized and the X-ray
image can be overlaid in a similar fashion to Navab et al.
\cite{navab2010camera}. A volumetric reconstruction of the scene is computed
using the RGBD data from the 2 cameras, following the principle of Truncated
Signed Distance Field (TSDF), used for example by Kinect Fusion
\cite{newcombe2011kinectfusion}. Then, the image is synthesized using raytracing
from the X-ray source viewpoint.
Knowing that the reconstruction is volumetric and that the 2 RGBD cameras are
positioned on the sides of the X-ray source, the cameras provide more
information than is actually used during raytracing. Indeed, the raytracing will
stop at the first voxel representing the surface (where the field is equal to
zero). If, instead of stopping at this voxel, the raytracing would go further
and search for the second voxel where the field is zero along the ray, a second
layer could be synthesized beyond the first layer. Thus, using a depth augmented
C-arm technology, this method would allow visualization of several layers. These
include front and back layers, which are equivalent to any instrument and
clinician hand above the patient anatomy, and the X-ray image plane
respectively.

Making the front layer transparent or even disappear in order to visualize what
is beyond has been studied in Diminished Reality (DR). In contrast to Augmented
Reality where graphics are overlaid to a real-scene, DR withdraws or attenuates
real elements from a scene. The works in DR can be divided into 3 categories
according to the background recovering method: multi-viewpoint, temporal, and
inpainting. The temporal methods
\cite{shen2006video,cosco2009augmented,buchmann2005interaction} suppose that the
camera have seen the scene without the occluder (or at another position) and use
this previous information to recover the current occluded pixels.
The inpainting methods recover the occluded part of an image with information
from its non-occluded part using patch-based methods
\cite{herling2010advanced,kawai2013diminished} or combined pixels methods
\cite{herling2012pixmix}.
The multi-viewpoint techniques use additional cameras that can observe the
occluded background totally, or partially in order to recover it from the
occluded viewpoint. Jarusirisawad and Saitoo \cite{Jarusirisawad2007diminsihed}
use perspective wrapping from the non-occluded cameras to the occluded camera to
recover background pixels. More recently, using RGBD cameras, several works
\cite{meerits2015real,saito2014camera} have generated surface mesh models of the
background from one or multiple side cameras. Observing the mesh from the
occluded viewpoint requires only a rigid transformation, avoiding distortions
due to wrapping. Sugimoto et al. \cite{sugimoto2014half} use the 3D geometry to
backproject to the side views the occluded pixels and therefore recover it.
By design, the multi-viewpoint recovery can be used  for the
stereo-RGBD augmented C-arm which contains 2 RGBD cameras are placed on the
side of the X-ray source viewpoint. Instead of using a mesh, the volumetric
field can be used.
However, no work in literature has used volumetric field such as TSDF to
recover background information to the best of our knowledge. Concerning the
visualization of the foreground of the front layer in combination with the back
layer, the most used technique is transparency
\cite{buchmann2005interaction,sugimoto2014half}. As explained by Livingston et
al. in their review of depth cues for ``X-ray'' vision augmented reality
\cite{livingston2013pursuit}, transparency is indeed the most natural depth cues
as it can be experienced in the real world with transparent objects.

\subsection{Contribution}
In this paper, we propose a mixed reality multi-layer visualization of the surgeon hands
and surgical instruments using a stereo-RGBD augmented C-arm fluoroscope. This
visualization consists of multiple layers which can be blended into one single
view along the line of sight of the surgeon while offering different output as
the blending values are chosen differently. The front layer synthesized from the
X-ray source viewpoint by the stereo-RGBD augmented C-arm contains the surgeon
hands and surgical instruments, the second layer is the background containing
the surgical target (i.e. also synthesized by our algorithm), while the last
layer is the X-ray image displaying the anatomy. As any layer can be blended to
the others, our visualization proposes, for example, to display transparent
hands on the background, on which the X-ray can also be blended. The blending
parameters can be chosen on the fly and according to preferences or
workflow steps.

In summary, this work presents the potential in positively impacting the
following areas:
\begin{itemize}
 \item User-adjustable multiple layer visualization for Medical Mixed Reality
       \vspace{-2mm}
 \item Improved training medical students and residents by visualizing multiple
       layers to better understand surgical instrument positioning and alignment,
       as opposed to visualizing the global scene using traditional augmented reality methods.
       \vspace{-2mm}
 \item First work in Medical Mixed Reality combining Diminished and Augmented Reality \vspace{-2mm}
 \item First use of volumetric field using TSDF for Diminished/Mixed
       Reality
\end{itemize}

\section{Methodology}

The setup, calibration methods, and image synthesization used in this paper have been previously published by \cite{habert2015posteraugmenting}.
In the interest of brevity, we will not describe the calibration steps but we will
thoroughly describe the synthesization process since it is vital to our Mixed Reality multi-layer  visualization contribution.

\subsection{Setup}
\begin{figure}[h] \centering
 \includegraphics[width=8.46cm]{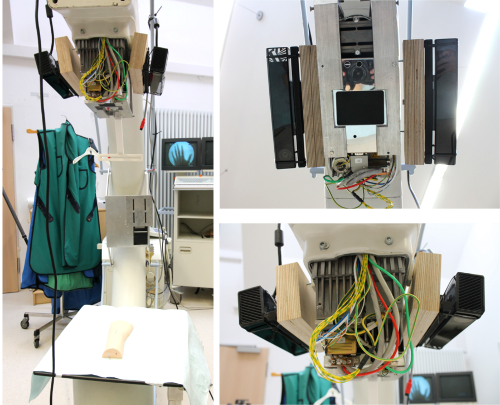}
 \caption{Setup with 2 Kinects attached on the C-arm gantry}
 \label{figsetup}
\end{figure}

The setup comprises 2 RGBD cameras  (Kinect v2) placed on the side of a X-ray
source \ref{figsetup}. Each RGBD camera outputs a depth image, an
infrared image and a wide-angle video image.
Their fields-of-view are overlapping over the C-arm detector.
Kinect v2 has been chosen because its depth information does not
interfere with a similar sensor.
The depth and video images are recorded using the libfreenect2 library
\cite{florian_echtler_2016_45314}. The mapping from depth to video image is provided by the library.  The synchronization between
images from the two cameras has been performed manually because two Kinect v2
can not used on a single standard computer and are therefore run on two separate
computers.
As a consequence, every sequence is recorded at a lower framerate than a
standard 30fps video.

\subsection{Image synthesization}
Once the system has been calibrated following the steps from \cite{habert2015posteraugmenting}, the video image from the X-ray viewpoint can be synthesized.
First, the origin of the 3D world coordinate space $\Omega_R \subset
\mathbb{R}^3$  is positioned at the center of the volumetric grid, around the
C-arm intensifier.
Knowing the poses of the two RGBD
cameras relative to the X-ray source, the projection matrices
$\Pi^1$ and $\Pi^2$ for the 2 RGBD sensors can be computed. The notations
relative to the cameras are defined as follows:  optical center of the
first camera $C^1$, its depth image $I_d^1$ and color image $I_c^1$
(respectively, in the second camera $C^2$, $I_d^2$ and $I_c^2$).

To render the color image from the X-ray source viewpoint,
a volumetric TSDF field $f_v:\Omega_R \longmapsto [-1,1] $ is created which
maps a 3D point $\textbf{x} \in \Omega_R$ to a truncated signed
distance value. This value is the weighted mean of
the truncated signed distance values $v^1(\textbf{x})$ and $v^2(\textbf{x})$
computed respectively in the 2 RGBD sensor cameras. Therefore, the field $f_v$ follows Equation \ref{eq:f}.
\begin{equation}
 f_v(\textbf{x})=\frac{w^1(\textbf{x})v^1(\textbf{x})+w^2(\textbf{x})v^2(\textbf{x})}{w^1(\textbf{x})+w^2(\textbf{x})}
 \label{eq:f}
\end{equation}
where $w^1$ and $w^2$ are the weights for each camera. The weights
are used to reject truncated signed values according to specific conditions
(described in Equation \ref{eq:w}).
For each camera $i\in \{1,2\}$, the weights $w^i(\textbf{x})$ for each truncated
signed value are computed as:
\begin{equation}
 w^i(\textbf{x})=
 \left\{
 \begin{array}{lr}
  1 \text{   if   } I_d^i(\Pi^i(\textbf{x}))-||\textbf{x}-C^i||<-\eta \\
  0  \text{    else}
 \end{array}
 \right.
 \label{eq:w}
\end{equation}
where $\eta$ is a tolerance on the visibility of $\textbf{x}$ (we use
$\eta=6mm$).
For each view $i\in \{1,2\}$,  $v^i(\textbf{x})$ represents geometrically the
difference in between the distance from $\textbf{x}$ to the optical center of the camera $i$ $C^i$  and the
depth value obtained by projecting $\textbf{x}$ into camera $i$, on which  a
scaled truncation to the interval [-1,1] is applied.
The truncated signed distances $v^i(\textbf{x})$
are computed according to Equation \ref{eq:v}.
\begin{equation}
 v^i(\textbf{x})=\phi(I_d^i(\Pi^i(\textbf{x}))-||\textbf{x}-C^i||) \text{ with
  } \phi(s)= \left\{
 \begin{array}{lr}
  sgn(s) \text{   if   } \frac{|s|}{\delta}>1 \\
  \frac{s}{\delta} \text{    else}
 \end{array}
 \right.
 \label{eq:v}
\end{equation}
with $\delta$ being a tolerance parameter to handle noise in depth measurements
($\delta=2mm$ in our method) .
Alongside with the TSDF $f_v$, we also create a volumetric color field $f_c:
\Omega_R \longmapsto [0..255]^3 $ following Equation \ref{eq:fc}.
\begin{equation}
 f_c(\textbf{x})=\frac{w^1(\textbf{x})I_c^1(\Pi^1(\textbf{x}))+w^2(\textbf{x})I_c^2(\Pi^2(\textbf{x}))}{w^1(\textbf{x})+w^2(\textbf{x})}
 \label{eq:fc}
\end{equation}

The scene to synthesize is represented in the volumetric grid by the voxels
whose TSDF values is equal to 0.
The color image $I_c$ from the X-ray viewpoint is therefore generated by
performing raytracing from the X-ray viewpoint on the TSDF field $f_v$.
For every pixel in the image to be synthesized, a ray is traced
passing through the X-ray source and the pixel.
Raytracing consists at searching the closest to the X-ray source voxel
$\textbf{y}$ respecting the condition $f_v(\textbf{y})=0$ along this ray.
To speed up this step, the search for the \mbox{0-value} is performed by binary
search.
Once the $\textbf{y}$ has been found, the color $f_c(y)$ is applied to the pixel
in the synthesized image $I_c$.
A depth image $I_d$ can be synthesized by calculating the distance between
$\textbf{y}$ and the X-ray source.

\subsection{Multi-Layer Image Generation}

\begin{figure}[h] \centering
 \includegraphics[width=5.50cm]{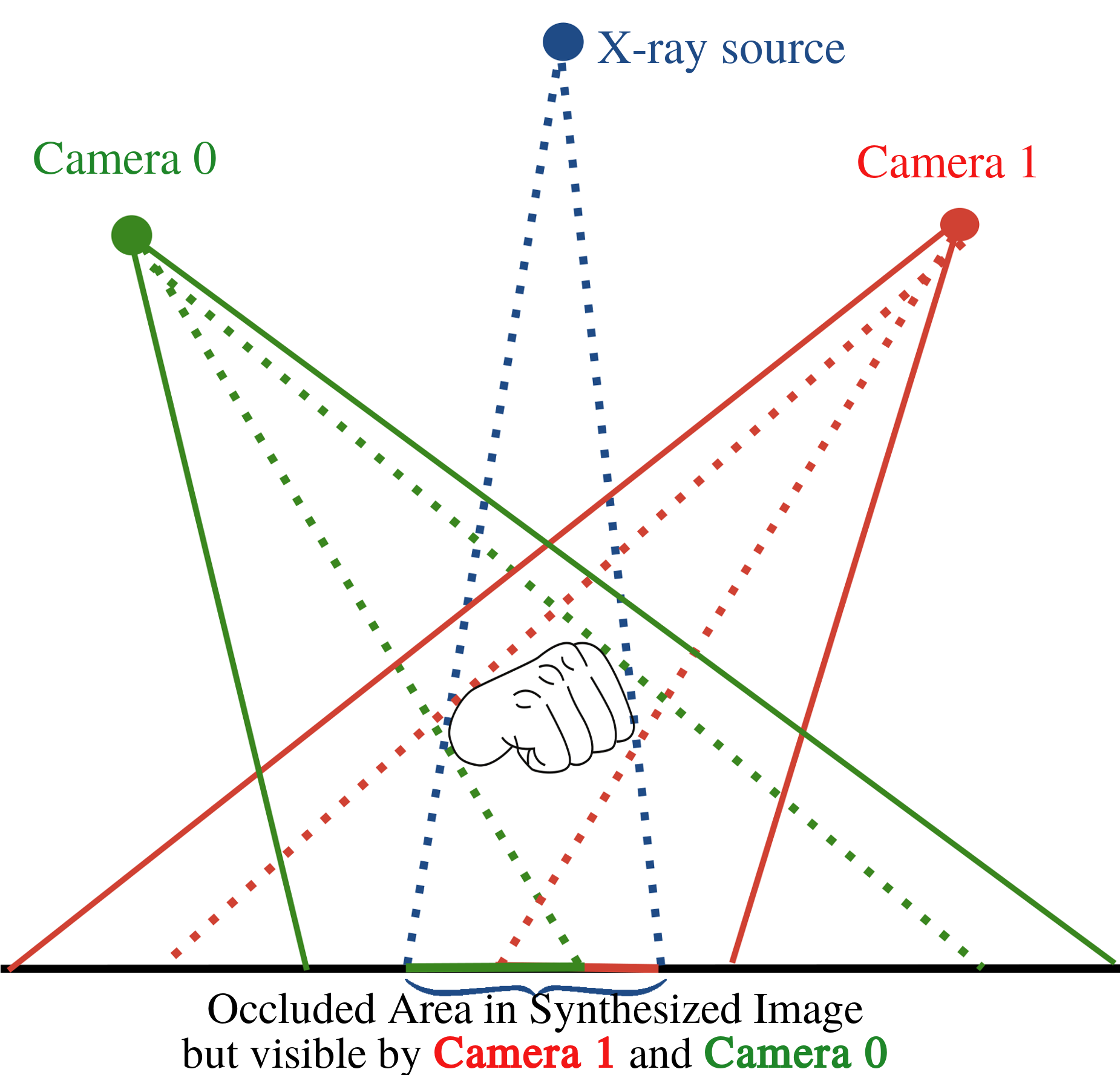}
 \caption{Occlusion}
 \label{occlusion}
\end{figure}

After the first raytracing step, the  video image $I_c$ as seen by
the X-ray source viewpoint, as well as its corresponding depth image
$I_d$ are generated.
The volumetric TSDF field is a dense representation which contains information
about the full 3D space around the C-arm detector whereas  the
raytracing stops only at the first found \mbox{\mbox{0-value}} voxel.
Therefore, the TSDF field contains more information than is actually used
until now.
Beyond the hands synthesized by the first raytracing, more \mbox{0-value} can be
present along the ray. This is
true especially since the 2 RGBD cameras are placed on the
side of the C-arm, giving additional information from another viewpoint.
This situation is illustrated in Figure \ref{occlusion} where the background
occluded by a hand from the X-ray source viewpoint (the blue point) can be seen
by at least one of the 2 cameras. In a TSDF representation, this means those occluded
background voxels also have a \mbox{0-value}.
To find those additional \mbox{0-value}, a modified ``second run'' raytracing must
be performed on the foreground (e.g. surgeon hands or surgical
tools).

\subsubsection{Hand segmentation}
\begin{figure}[h]
 \centering
 \begin{subfigure}[b]{0.22\textwidth}
  \includegraphics[width=\textwidth]{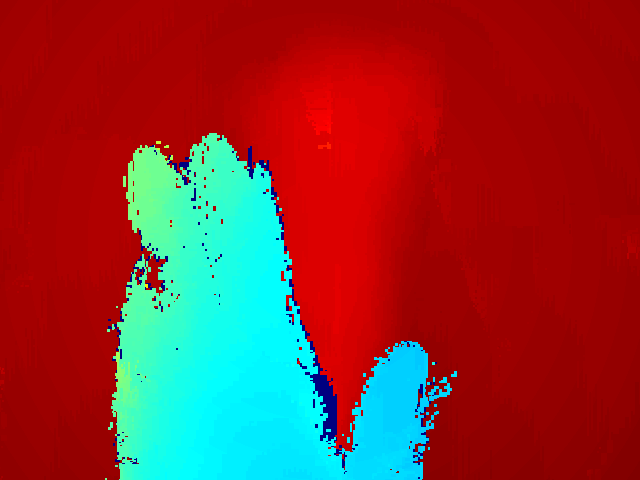}

  \label{fig:video0}
 \end{subfigure}
 \begin{subfigure}[b]{0.22\textwidth}
  \includegraphics[width=\textwidth]{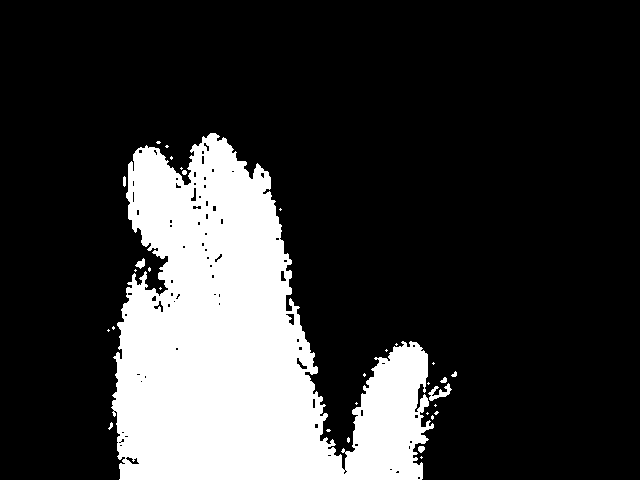}

  \label{fig:video0}
 \end{subfigure}
 \caption{The synthesized depth image and its corresponding
  segmented mask\label{mask}}
\end{figure}

As a first step, the foreground needs to be segmented from the synthesized video
image $I_c$ and depth image $I_d$.
A background model is computed from an initialization sequence of depth images
where no hands or surgical instruments are introduced yet.
An average depth image is created by averaging the depth at every pixel along
the initialization sequence.
Then, for every new image (with potential hands or surgical instruments present), the depth image $I_d$
is compared to the mean image in order to create a binary mask image $I_m$.
For every pixel whose depth is lower than the average depth minus a margin
(3 cm), the pixel is classified as foreground and is set as white in $I_m$.
If the pixel is classified as background, then it is set as black in $I_m$.
The method is rudimentary compared to background subtraction methods, however
the margin allows the background to change shape (in the limit of the margin).
A noise removal step is added using morphological opening on the mask image. An
example of scaled depth image and its corresponding mask are shown on Figure
\ref{mask}.

\subsubsection{Second-run raytracing}

Once the foreground has been segmented, a second raytracing can be performed on
the pixels classified as hands or surgical instruments.
Instead of beginning the raytracing from the X-ray source viewpoint, the ray
search starts at the voxel $\textbf{y}$ found at the first
raytracing run plus a margin of 4 cm. This margin is the insurance to not find
a \mbox{0-value} still related to the foreground. The starting voxel $\textbf{y}$ can be
easily retrieved using the depth image $I_d$ resulting from the first
raytracing.
The raytracing is then performed forward using binary search in a similar fashion to
the first run of raytracing.
As a result, a color image of the background can be synthesized and
combined to the color image from the first raytracing run (excluding the
foreground segmented pixels) creating a complete background image $I_b$.

\subsubsection{Multi-Layer Visualization}

On top of the background image $I_b$, the foreground layer extracted from $I_c$
can be overlaid  with transparency as well as the X-ray image $I_{xray}$.
A multi-layer image $I_{layers}$ can then be created by blending all the layers
according to Equation \ref{eq:beta1}.
\begin{equation}
 I_{layers}(p)=
 \left\{
 \begin{array}{lr}
  \alpha I_{c}(p) +\beta I_b(p) +\gamma I_{xray}(p) \text{ if } p \in
  \text{foreground}                                                   \\
  (1-\delta) I_b(p) +\delta I_{xray}(p) \text{    else}
 \end{array}
 \right.
 \label{eq:beta1}
\end{equation}
where $(\alpha,\beta,\gamma,\delta) \in [0,1]^4$ with $\alpha+\beta+\gamma=1$
are the blending parameters associated with each level. They can also be seen as
specific weight values which emphasize a specific layer during the blending process.

%

\begin{figure}[h] \centering
 \includegraphics[width=5cm]{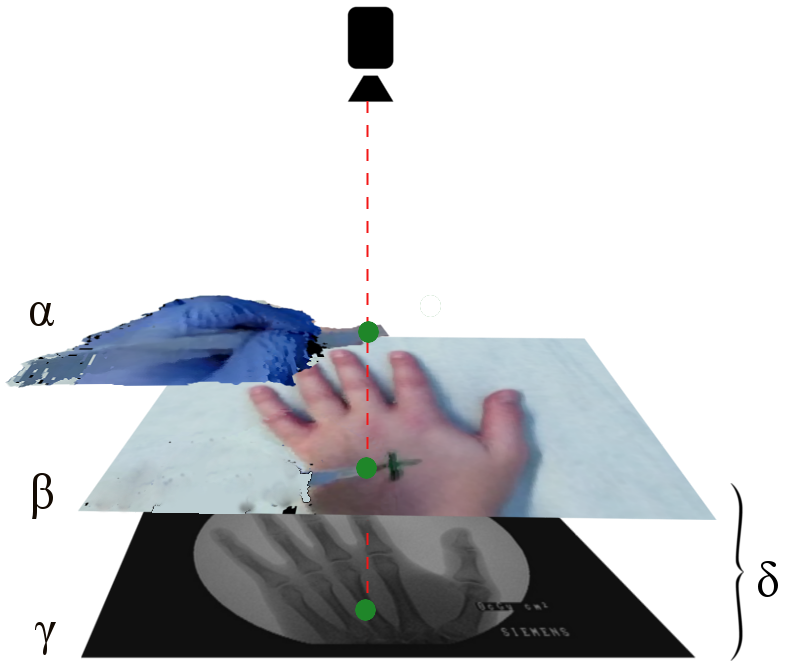}
 \caption{Layers in our visualization, all can be observed depending on
  the chosen blending values $\alpha,\beta,\gamma,\delta$ }
 \label{layers}
\end{figure}

The visualization scheme we propose allows us then to
observe three layers of structures (displayed in Figure \ref{layers}) according
to those parameters.

The furthest layer is the X-ray, which can be observed in its
totality in the image $I_{layers}$ with
$(\alpha,\beta,\gamma,\delta)=(0,0,1,1)$.
As we get closer to the camera, another layer is the background structure recovered using volumetric field. It can be observed with
$(\alpha,\beta,\gamma,\delta)=(0,1,0,0)$.
Finally the front layer comprising the hands  and instruments can be observed in
the image $I_{layers}$ using $(\alpha,\beta,\gamma,\delta)=(1,0,0,0)$.
Our visualization scheme allows to see in transparency the different layers
(anatomy by X-ray, background, front layer ) by choosing blending parameters
$(\alpha,\beta,\gamma,\delta)$ non equal to 0 and 1.
The choice of blending values depends on multiple parameters such as surgeon
preferences, step in the surgical workflow, type of instrument used. It can be changed
on the fly during surgery according to such parameters.
For example, once an instrument has already penetrated the skin, the background
is not necessary to visualize. The transparent hands can be overlaid directly on
the X-ray image, skipping the background layer. This scenario corresponds to
blending parameters $(\beta,\delta)=(0,1)$, $\alpha=1-\gamma$ with $0 <\gamma <
1$. With the configuration $(\alpha,\beta,\gamma,\delta)=(1,0,0,1)$, the
visualization consists of fully opaque hands or surgical tools on the X-ray
image, giving a similar output as \cite{pauly2014relevance} which aimed at
obtaining a natural ordering of hands over X-ray image.
%
%
As every layer is known
at any point in a sequence, the multi-layer visualization can be replayed to
medical students and residents for example with other blending parameters than
the one used in surgery.
They can have full control for the observation of the layers having
the choice to emphasize particular layers of interest for their learning.

\section{Results}

%

\subsection{Experimental protocol}

Six sequences have been recorded depicting example scenarios which include both
surgeon hands and surgical tools. Both a realistic hand model phantom and a real
patient hand are used and positioned on a surgical table. A clinician wearing
purple examination gloves introduces partial occlusions randomly to the scene.
Sequences 1 and 3 contain the motion of the clinician’s hand above the hand
model phantom at 20 cm and 30 cm respectively. Sequences 2 and 4 contain the
motion of a clinician’s hand closed and above the hand model phantom at 20 cm
and 30 cm respectively. Sequences 3 and 4 also contain incision lines drawn
using a marker on the hand model phantom. Finally, Sequences 5 and 6 are
recorded with surgical tools above a real patient hand. Sequence 5 includes
actions using a surgical hammer aiming for a cross target drawn on the patient
hand. Sequence 6 includes a scalpel targeting the same cross. The
heights of the surgical instruments to the patient hand vary up from 5 cm to 30
cm.

\subsection{Background recovery}
\label{back_rec}

For every sequence,  the mean value for the percentage of recovered pixels is
calculated and indicated in Table \ref{tableerrorproj}. The natural observation
in Table \ref{tableerrorproj} is that the closer the surgeon hand and surgical tools are to the anatomy the larger the occlusion in both side cameras
will be.
This signifies a lower percentage of recovered pixels by our algorithm which is demonstrated.

\begin{table}[h]
 \begin{center}
  \begin{adjustbox}{max width=8.46 cm}
   \begin{tabular}{ c c c c c c c}
    \toprule
    Sequences                & 1    & 2    & 3    & 4    & 5    & 6    \\
    \midrule
    Pixels recovered (in \%) & 69.3 & 65.2 & 88.2 & 97.4 & 84.1 & 45.2 \\
    \bottomrule
   \end{tabular}
  \end{adjustbox}
 \end{center}
 \vspace{-0.3cm}
 \caption{Background recovery results}
 \label{tableerrorproj}
\end{table}

Sequences 1 and 2 were recorded with surgeon hand open (69.3\%) and closed
(65.2\%)
Less pixels are recovered for the close hand scenario as mainly the fist is
present in the scene. The fist is also not recovered  in the other scenario but
the fingers are also occluding which are easier to recover from (due
to their thin shape), in percentage, the open hand scenario recovers more, even
if occluding more.
Sequences 3 and 4 resulted in larger recovery percentages (88.2\% and 97.4\%
respectively) because the surgeon hand was farther away from the hand model.
This implies that there is a greater probability for the background
voxels to be seen by the RGBD sensors.
Sequence 6 with a scalpel confirms that the height strongly influences the
recovery. The scalpel scenario which includes numerous images with hands and
instruments close to the background (less than 10 cm) shows a low recovery
result as expected.
Due to the hammer's shape, the sequence 5 shows however a higher recovery
percentage.

\subsection{Visualization results}

\begin{figure*}
 \begin{turn}{90}
  \begin{minipage}{21.5cm}
   \centering
   \begin{subfigure}[t]{0.163\textwidth}
    \includegraphics[width=\textwidth]{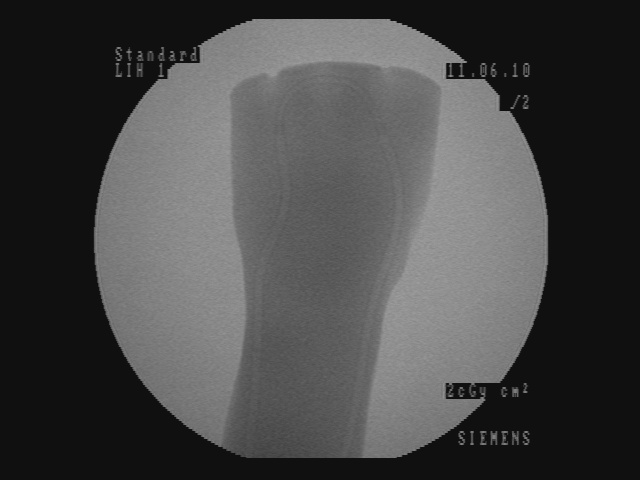}
    \label{fig:video0}
   \end{subfigure}
   \begin{subfigure}[t]{0.163\textwidth}
    \includegraphics[width=\textwidth]{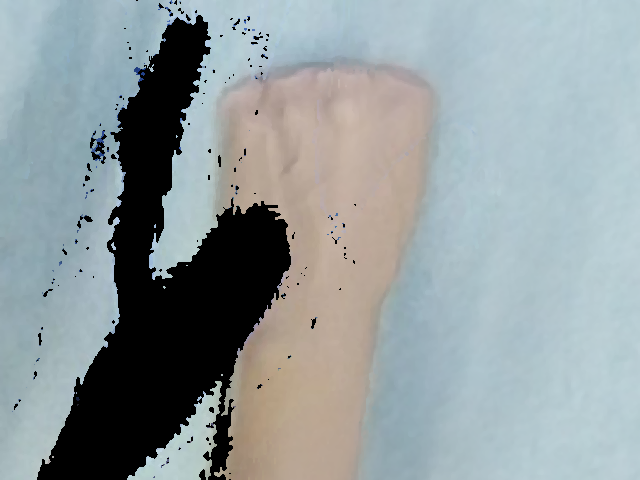}
    \label{fig:video0}
   \end{subfigure}
   \begin{subfigure}[t]{0.163\textwidth}
    \includegraphics[width=\textwidth]{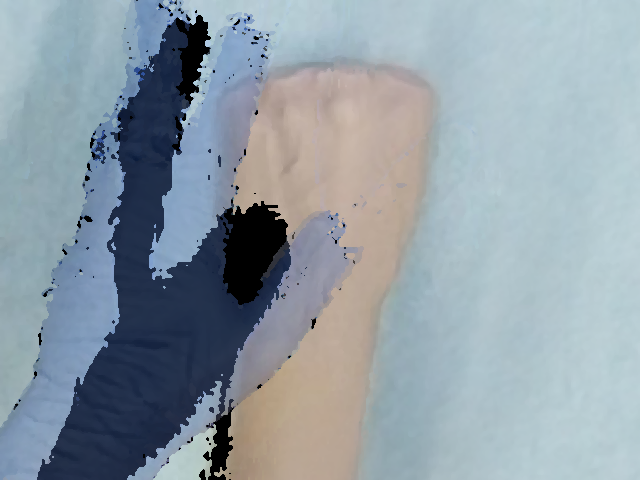}
    \label{fig:xray0}
   \end{subfigure}
   \begin{subfigure}[t]{0.163\textwidth}
    \includegraphics[width=\textwidth]{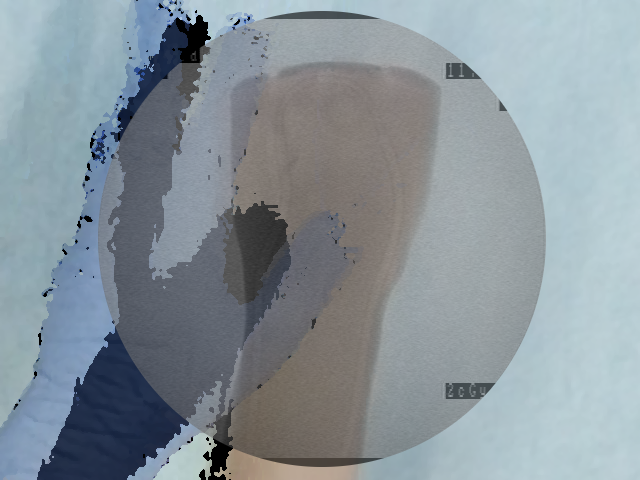}
    \label{fig:xray0}
   \end{subfigure}
   \begin{subfigure}[t]{0.163\textwidth}
    \includegraphics[width=\textwidth]{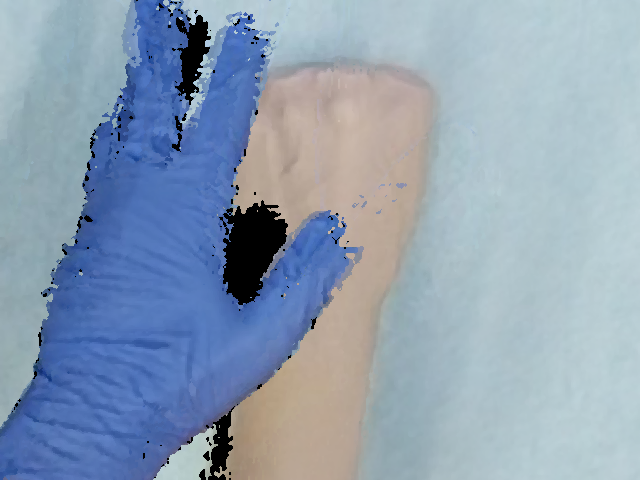}
    \label{fig:xray0}
   \end{subfigure}
   \begin{subfigure}[t]{0.163\textwidth}
    \includegraphics[width=\textwidth]{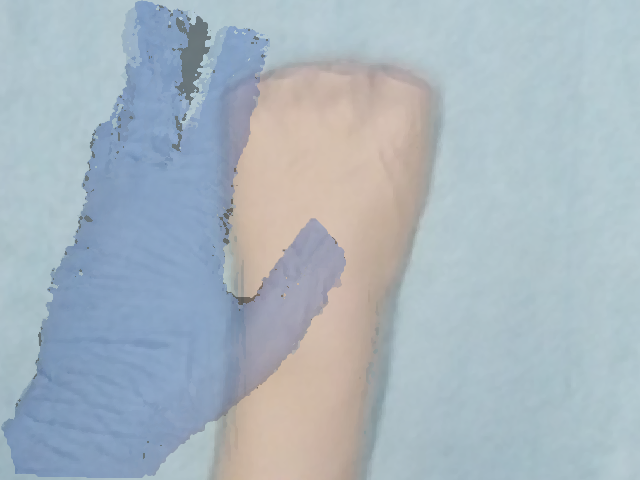}
    \label{fig:xray0}
   \end{subfigure}

   \begin{subfigure}[t]{0.163\textwidth}
    \includegraphics[width=\textwidth]{xrayseq1.jpg}
    \label{fig:video0}
   \end{subfigure}
   \begin{subfigure}[t]{0.163\textwidth}
    \includegraphics[width=\textwidth]{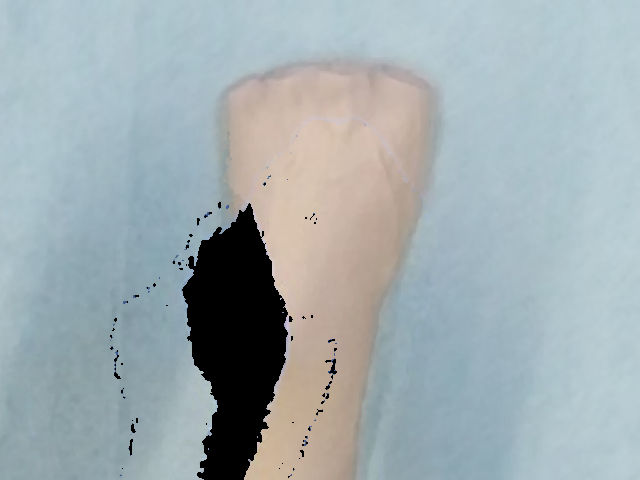}
    \label{fig:video0}
   \end{subfigure}
   \begin{subfigure}[t]{0.163\textwidth}
    \includegraphics[width=\textwidth]{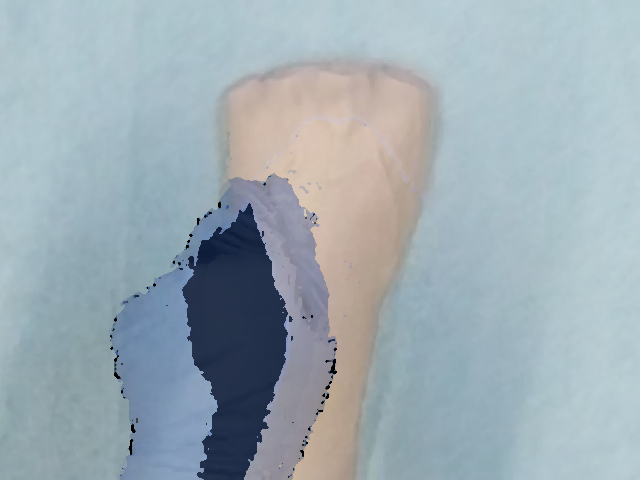}
    \label{fig:xray0}
   \end{subfigure}
   \begin{subfigure}[t]{0.163\textwidth}
    \includegraphics[width=\textwidth]{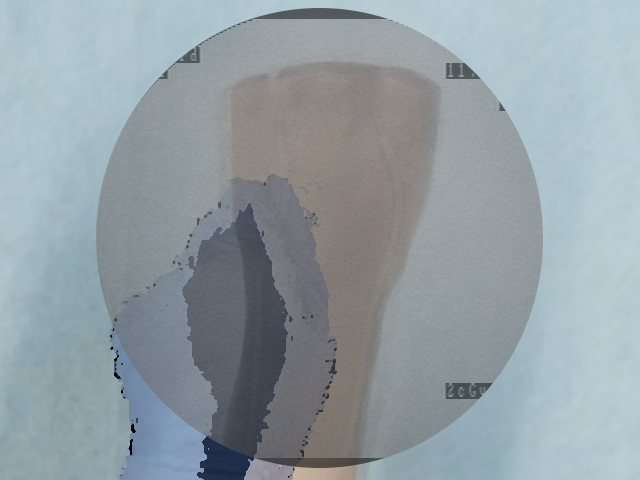}
    \label{fig:xray0}
   \end{subfigure}
   \begin{subfigure}[t]{0.163\textwidth}
    \includegraphics[width=\textwidth]{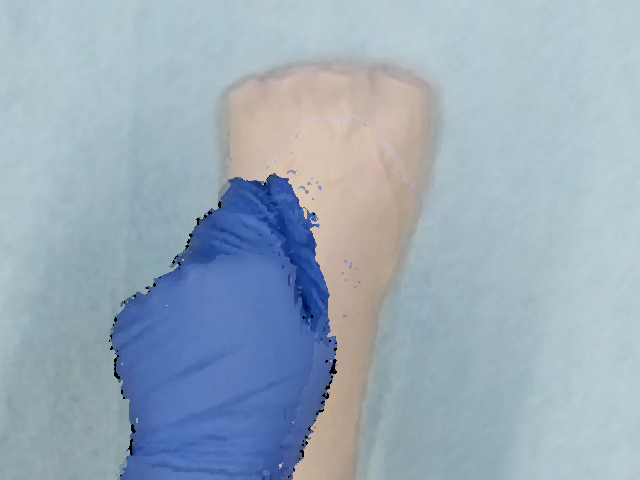}
    \label{fig:xray0}
   \end{subfigure}
   \begin{subfigure}[t]{0.163\textwidth}
    \includegraphics[width=\textwidth]{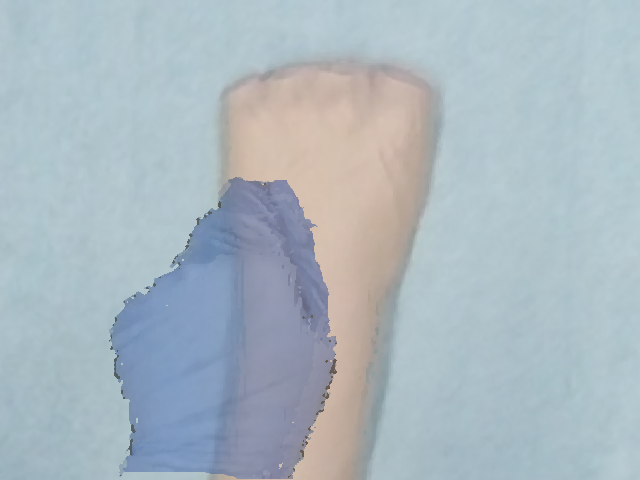}
    \label{fig:xray0}
   \end{subfigure}

   \begin{subfigure}[t]{0.163\textwidth}
    \includegraphics[width=\textwidth]{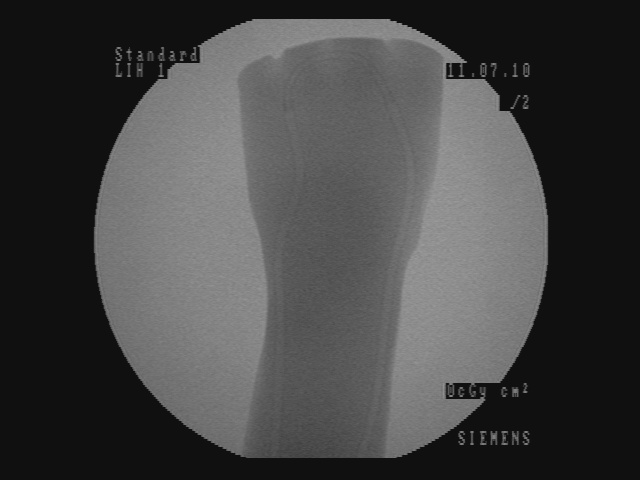}
    \label{fig:video0}
   \end{subfigure}
   \begin{subfigure}[t]{0.163\textwidth}
    \includegraphics[width=\textwidth]{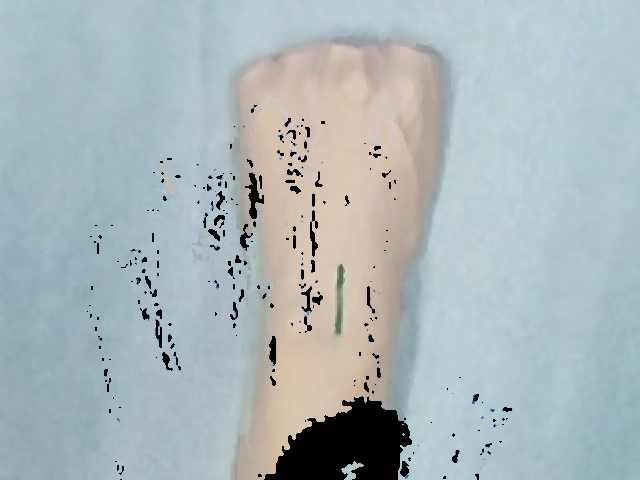}
    \label{fig:video0}
   \end{subfigure}
   \begin{subfigure}[t]{0.163\textwidth}
    \includegraphics[width=\textwidth]{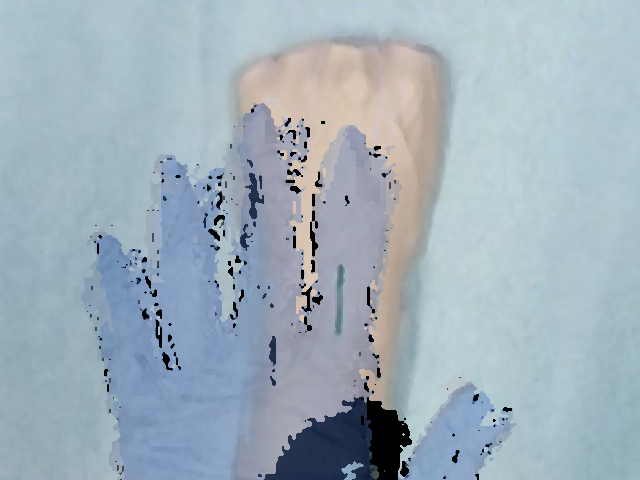}
    \label{fig:xray0}
   \end{subfigure}
   \begin{subfigure}[t]{0.163\textwidth}
    \includegraphics[width=\textwidth]{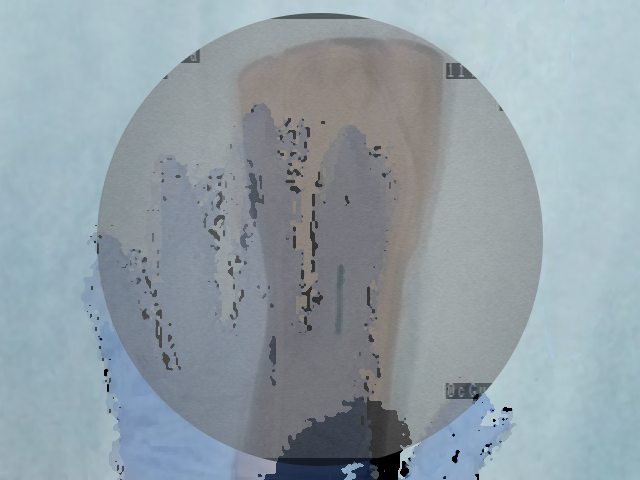}
    \label{fig:xray0}
   \end{subfigure}
   \begin{subfigure}[t]{0.163\textwidth}
    \includegraphics[width=\textwidth]{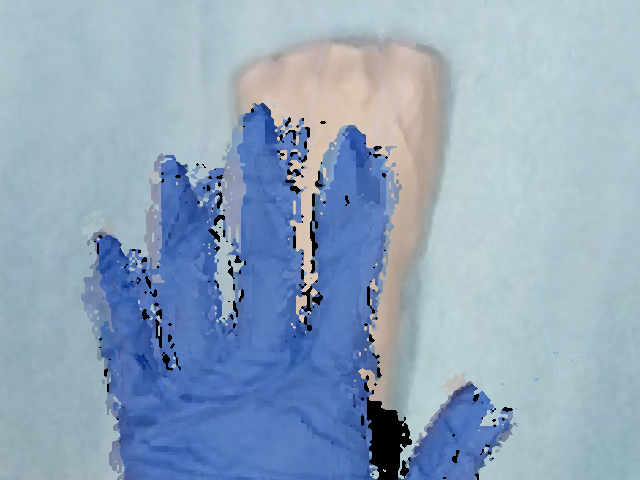}
    \label{fig:xray0}
   \end{subfigure}
   \begin{subfigure}[t]{0.163\textwidth}
    \includegraphics[width=\textwidth]{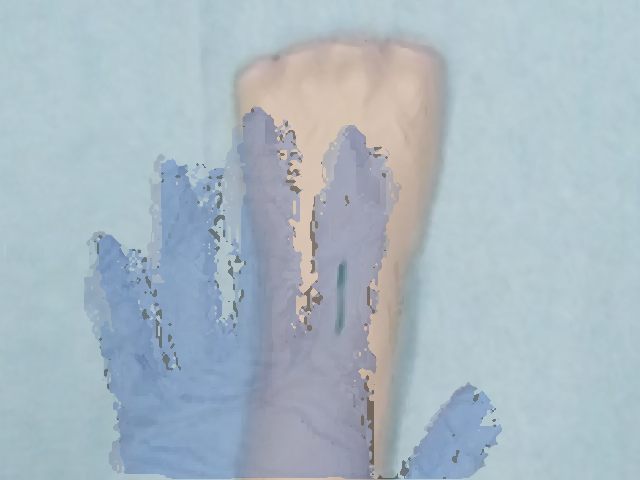}
    \label{fig:xray0}
   \end{subfigure}
   \begin{subfigure}[t]{0.163\textwidth}
    \includegraphics[width=\textwidth]{xrayseq2.jpg}
    \label{fig:video0}
   \end{subfigure}
   \begin{subfigure}[t]{0.163\textwidth}
    \includegraphics[width=\textwidth]{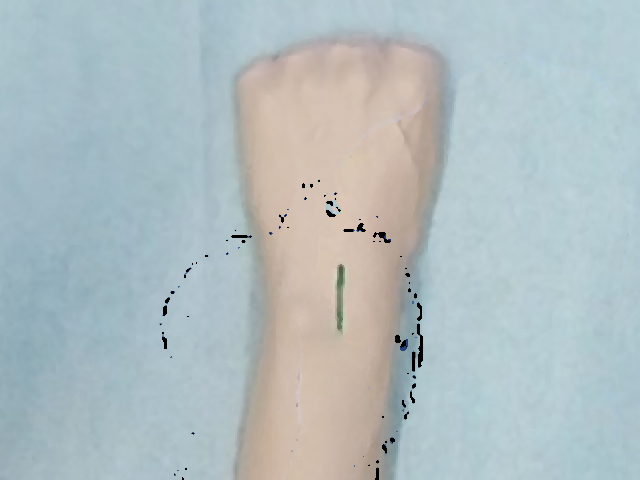}
    \label{fig:video0}
   \end{subfigure}
   \begin{subfigure}[t]{0.163\textwidth}
    \includegraphics[width=\textwidth]{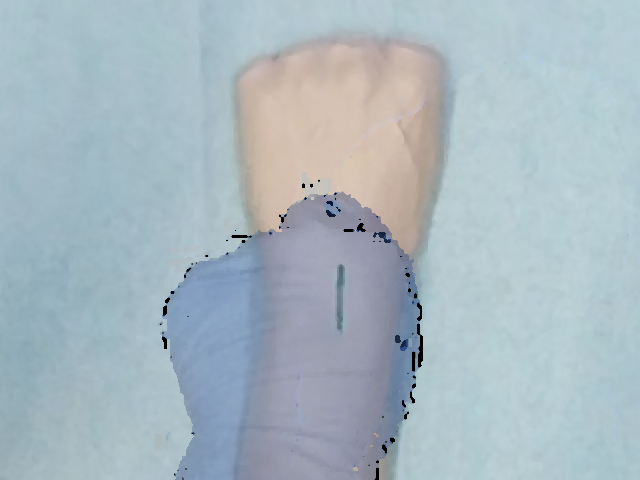}
    \label{fig:xray0}
   \end{subfigure}
   \begin{subfigure}[t]{0.163\textwidth}
    \includegraphics[width=\textwidth]{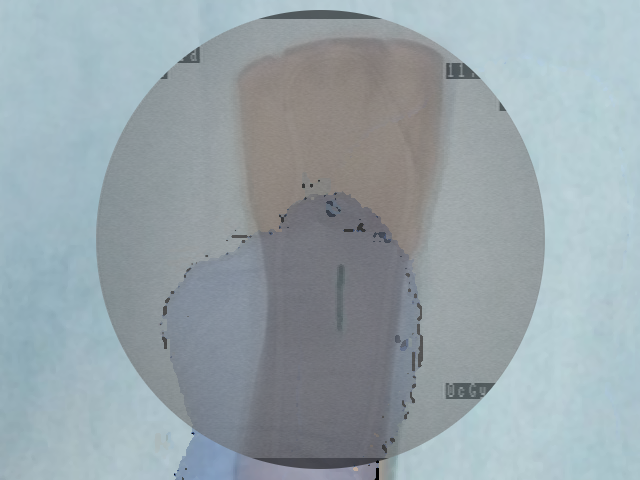}
    \label{fig:xray0}
   \end{subfigure}
   \begin{subfigure}[t]{0.163\textwidth}
    \includegraphics[width=\textwidth]{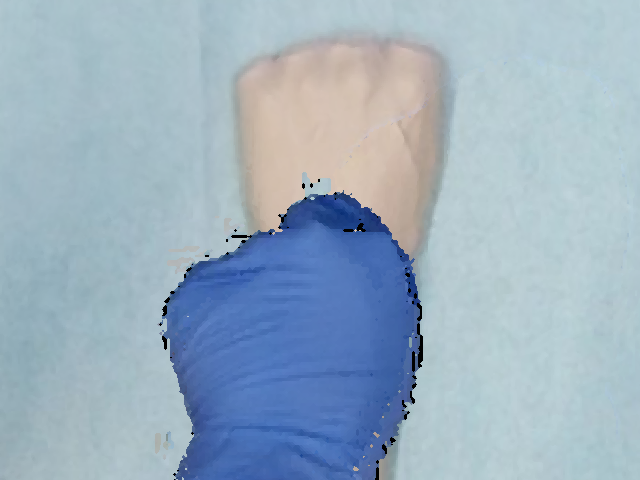}
    \label{fig:xray0}
   \end{subfigure}
   \begin{subfigure}[t]{0.163\textwidth}
    \includegraphics[width=\textwidth]{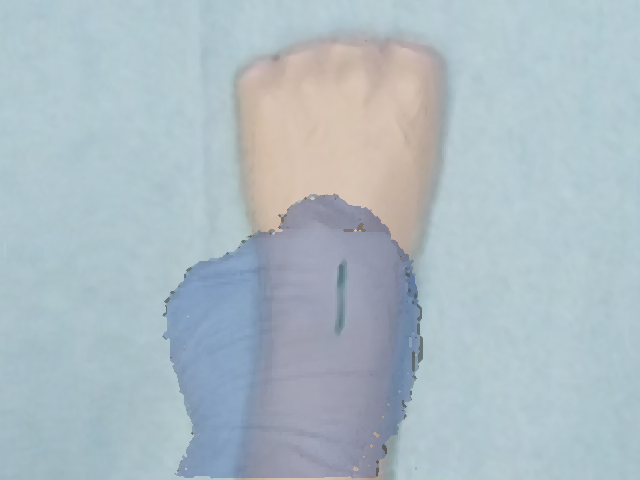}
    \label{fig:xray0}
   \end{subfigure}
   \begin{subfigure}[t]{0.163\textwidth}
    \includegraphics[width=\textwidth]{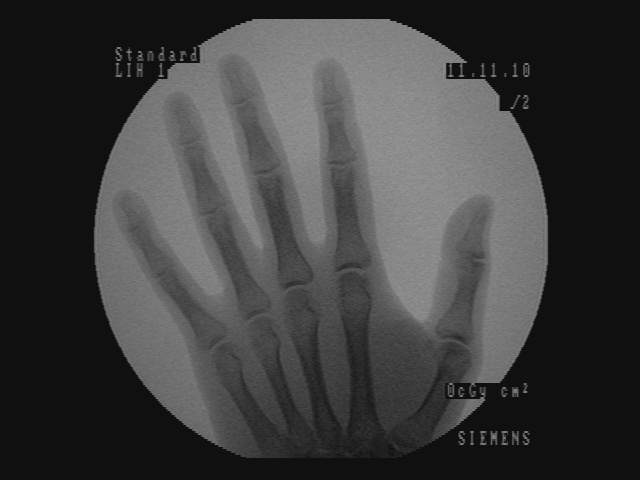}
    \label{fig:video0}
   \end{subfigure}
   \begin{subfigure}[t]{0.163\textwidth}
    \includegraphics[width=\textwidth]{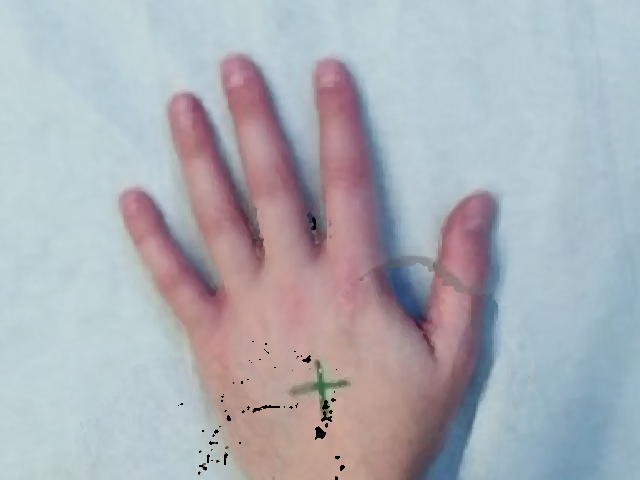}
    \label{fig:video0}
   \end{subfigure}
   \begin{subfigure}[t]{0.163\textwidth}
    \includegraphics[width=\textwidth]{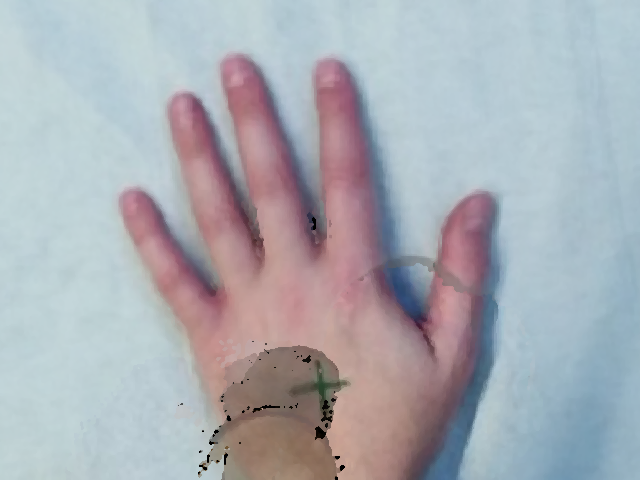}
    \label{fig:xray0}
   \end{subfigure}
   \begin{subfigure}[t]{0.163\textwidth}
    \includegraphics[width=\textwidth]{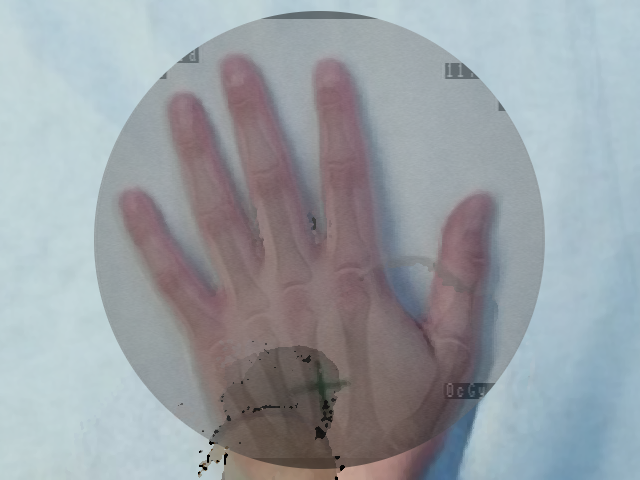}
    \label{fig:xray0}
   \end{subfigure}
   \begin{subfigure}[t]{0.163\textwidth}
    \includegraphics[width=\textwidth]{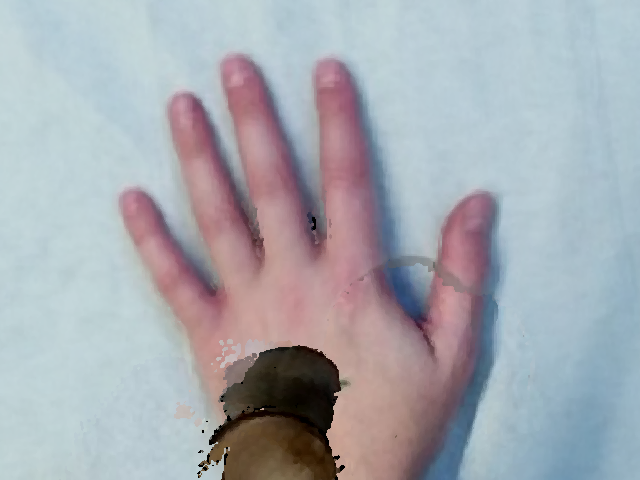}
    \label{fig:xray0}
   \end{subfigure}
   \begin{subfigure}[t]{0.163\textwidth}
    \includegraphics[width=\textwidth]{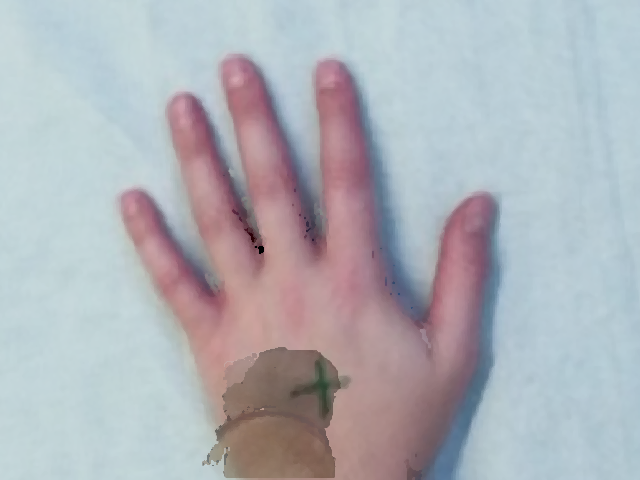}
    \label{fig:xray0}
   \end{subfigure}
   \begin{subfigure}[t]{0.163\textwidth}
    \includegraphics[width=\textwidth]{xray1.jpg}
    \subcaption{ $(\alpha,\beta,\gamma,\delta)=(0,0,1,1)$}
    \label{fig:video0}
   \end{subfigure}
   \begin{subfigure}[t]{0.163\textwidth}
    \includegraphics[width=\textwidth]{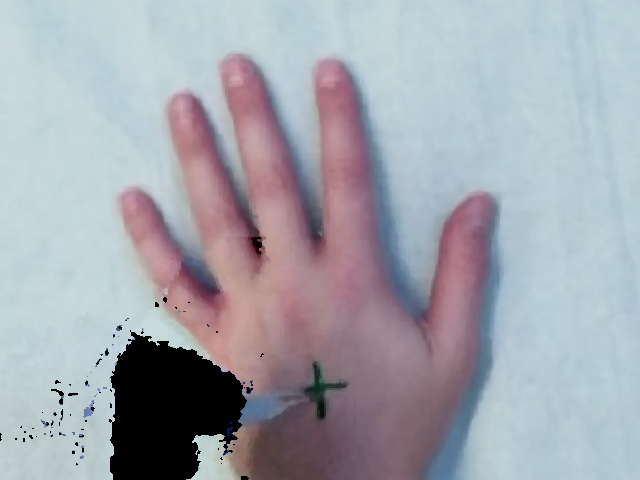}
    \subcaption{$(\alpha,\beta,\gamma,\delta)=(0,1,0,0)$}
    \label{fig:video0}
   \end{subfigure}
   \begin{subfigure}[t]{0.163\textwidth}
    \includegraphics[width=\textwidth]{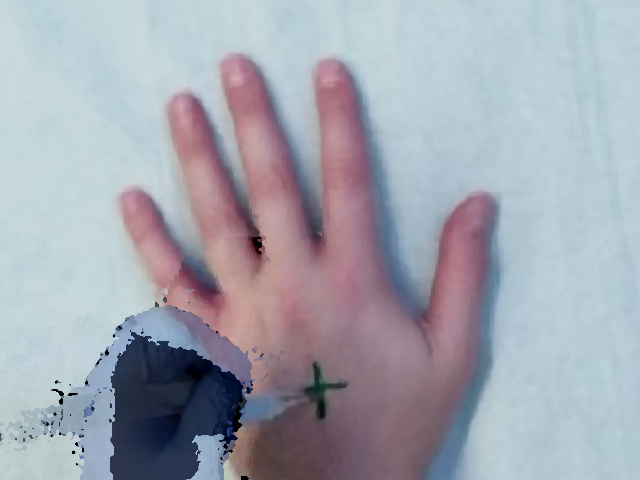}
    \subcaption{$(\alpha,\beta,\gamma,\delta)=(0.4,0.6,0,0)$}
    \label{fig:xray0}
   \end{subfigure}
   \begin{subfigure}[t]{0.163\textwidth}
    \includegraphics[width=\textwidth]{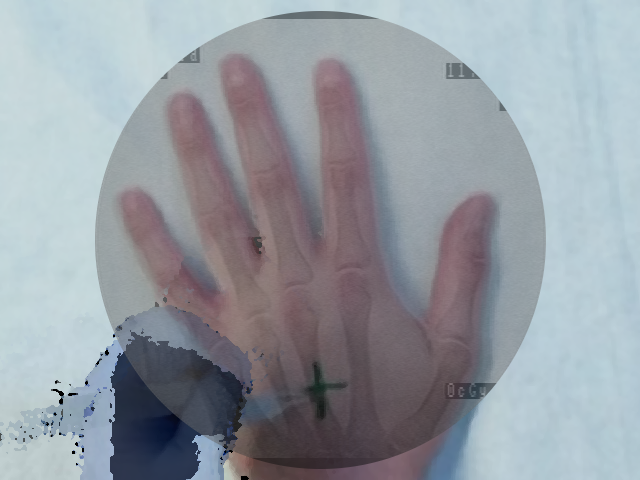}
    \subcaption{$(\alpha,\beta,\gamma,\delta)=(0.2,0.3,0.5,0.5)$}
    \label{fig:xray0}
   \end{subfigure}
   \begin{subfigure}[t]{0.163\textwidth}
    \includegraphics[width=\textwidth]{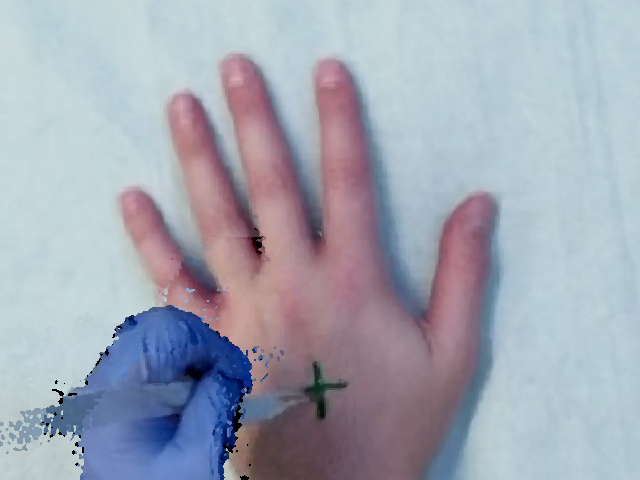}
    \subcaption{$(\alpha,\beta,\gamma,\delta)=(1,0,0,0)$}
    \label{fig:xray0}
   \end{subfigure}
   \begin{subfigure}[t]{0.163\textwidth}
    \includegraphics[width=\textwidth]{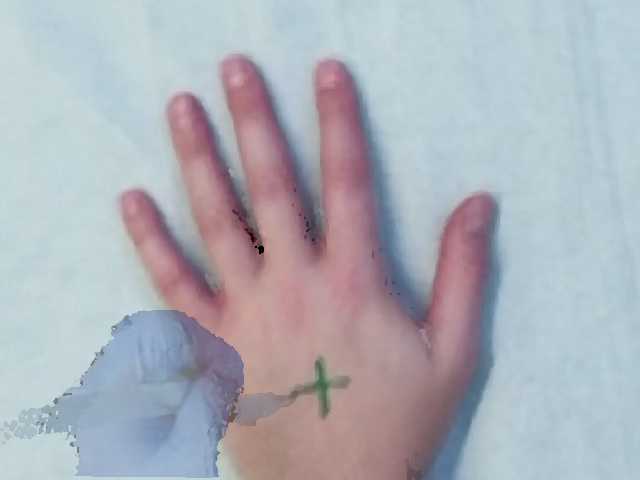}
    \subcaption{Inpainting}
    \label{fig:xray0}
   \end{subfigure}
  \end{minipage}
 \end{turn}
 \caption{Per row $i$, multi-layer image $I_{layers}$ of one selected frame in
  the sequence $i$ with different blending parameters
  $(\alpha,\beta,\gamma,\delta)$
  \label{goodrenderingresults}}
\end{figure*}

In Figure \ref{goodrenderingresults}, for each scenario, one selected image
$I_{layers}$ in the sequence can observed with different values of $\alpha$,
$\beta$, $\gamma$ and $\delta$.
Each row $i$ corresponds to the sequence $i$. From left to right, the layer
visualized in $I_{layers}$ is getting closer to the X-ray source viewpoint. In
the column (a), the furthest layer (the X-ray image) is displayed. In the
column (b), the second layer (the background), in the column (c), the
blending of the front layer with the background, in the column (d), the
blending of the three layers and finally, in the column (e), the closest layer
is shown. Additional images from the sequences can be visualized in the
supplementary video where interaction between the layers by changing the
blending values can be observed.

Despite the fact that the background cannot be ideally recovered, a manual post
processing step involving inpainting is applied and displayed in the column (f)
of Figure \ref{goodrenderingresults}. We believe that the multi-layer
visualization concept is an interesting and profound solution offering numerous
possibilities in the surgical areas, as well as, the mixed reality communities.

Similar to results from Habert et al. \cite{habert2015posteraugmenting}, the
images resulting from synthesization are not as sharp as a real video image. The
area synthesized by our algorithm is approximately 20 cm $\times$ 20 cm (C-arm detector size), which
is small compared to the wide-angle field of view from the Kinect v2. Reduced
to the area of synthesization, the video and depth from Kinect is not of high
resolution enough for sharper results. More specialized hardware with smaller
field of view and higher resolution RGBD data would solve this problem.
Moreover, several artifacts can be seen around the hand and surgical
instruments in the synthesized image due to high difference and noise in depth
in the RGBD data from the 2 cameras.
However, our results demonstrate that our method is working well,
since the incision line and cross drawn on the hand model and patient hand
are perfectly visible in the recovered background image and can be seen in
transparency through the hands and surgical tools in the images of Figure
\ref{goodrenderingresults}-column (c) and (d).
In the scalpel sequence (sequence 6) in Figure
\ref{goodrenderingresults}-column (b), it can be seen that
the tip of the scalpel is considered as background, this is due to the margin of
few centimeters used for background segmentation. In this image, the scalpel is
actually touching the skin.

\section{Discussion}

Inferring temporal priors can help alleviate occlusion. Methods involving volumetric fields
\cite{newcombe2011kinectfusion} use temporal information as the field is
sequentially updating with new information, instead of fully being reinitialized as per our method.
The percentage of pixels recovered is also dependent of the side cameras configuration.
In our clinical case, the camera setup is constrained by the C-arm design and
the disparity between the X-ray source and the two RGBD cameras is low. A higher
disparity would lead to less occlusion in at least one of the cameras. Even with our
constrained and difficult clinical setup, the results are extremely promising
and we are convinced the work could also be easily extended to less restrictive
settings. A potential application is Industrial Diminished/Mediative Reality
where workers wearing a HMD with two cameras placed on its side (with a higher
disparity than our setup) could see their viewpoint synthesized with their hands
in transparency.

\section{Conclusion}

In this paper, we have presented the first work combining Diminished and Augmented Reality in
medical domain. Our visualization scheme proposes a user-adjustable multiple
layer visualization where each layer can be blended with others. The multiple
layers comprise the anatomy with the X-ray image, the patient background, and
the surgeon hand and surgical instruments. The result of our visualization
scheme offers the clinician to choose which layer(s) are to become transparent
depending on the surgical scenario or workflow step. Beyond the medical domain,
this work is the first use of volumetric field for background recovery in
Diminished Reality and Mixed Reality. Future works should involve adding
additional layers, by disassociating the surgeon hand layer from the surgical
instruments layer, in order to adjust further the visualization to the user preferences.

\bibliographystyle{abbrv}
\bibliography{biblio}

\end{document}